\documentclass[10pt,twocolumn,letterpaper]{article}

\usepackage{iccv}
\usepackage{times}
\usepackage{epsfig}
\usepackage{graphicx}
\usepackage{amsmath}
\usepackage{amssymb}
\usepackage{times}
\DeclareMathOperator*{\argmax}{arg\,max}
\DeclareMathOperator*{\argmin}{arg\,min}
\usepackage{rotating}
\usepackage{multirow}
\usepackage{tabulary}
\usepackage{float}
\usepackage{textcomp}
\usepackage[font=small,skip=0pt]{caption}
\usepackage[T1]{fontenc}
\newcommand{\ignore}[1]{}

\usepackage[pagebackref=true,breaklinks=true,letterpaper=true,colorlinks,bookmarks=false]{hyperref}

\iccvfinalcopy %

\def\iccvPaperID{5129} %

\ificcvfinal\fi
\begin{document}

\title{Fooling Network Interpretation in Image Classification}
\author{
  Akshayvarun Subramanya\footnotemark[1] \qquad
  Vipin Pillai\thanks{Equal contribution} \qquad  Hamed Pirsiavash\\
  University of Maryland, Baltimore County \\
  \texttt{\{akshayv1, vp7, hpirsiav\}@umbc.edu} \\
}

\maketitle

\begin{abstract}

Deep neural networks have been shown to be fooled rather easily using adversarial attack algorithms.
Practical methods such as adversarial patches have been shown to be extremely effective in causing misclassification. However, these patches are highlighted using standard network interpretation algorithms, thus revealing the identity of the adversary. We show that it is possible to create adversarial patches which not only fool the prediction, but also change what we interpret regarding the cause of the prediction. Moreover, we introduce our attack as a controlled setting to measure the accuracy of interpretation algorithms. We show this using extensive experiments for Grad-CAM interpretation that transfers to occluding patch interpretation as well. We believe our algorithms can facilitate developing more robust network interpretation tools that truly explain the network's underlying decision making process.

\end{abstract}

\section{Introduction}
 Deep learning has achieved great results in many domains including computer vision. However, it is still far from being deployed in many real-world applications due to reasons including:

\noindent {\bf (1) Explainable AI (XAI):} Explaining the prediction of deep neural networks is a challenging task because they are complex models with large number of parameters. Recently, XAI has become a trending research area in which the goal is to develop reliable interpretation algorithms that explain the underlying decision making process. Designing such algorithms is a challenging task and considerable work \cite{simonyan2013deep,zhou2016learning,selvaraju2016grad} has been done to describe \textit{local explanations} - explaining the model's output for a given input \cite{baehrens2010explain}. %

\noindent {\bf (2) Adversarial examples:} It has been shown that deep neural networks are vulnerable to adversarial examples. These carefully constructed samples are created by adding imperceptible perturbations to the original input for changing the final decision of the network. This is important for two reasons: (a) Such vulnerabilities could be used by adversaries to fool AI algorithms when they are deployed in real-world applications such as Internet of Things (IoT) \cite{mosenia2017comprehensive} or self-driving cars \cite{sitawarin2018darts} (b) Studying these attacks can lead to better understanding of how deep neural networks work and also possibly better generalization.

\begin{figure*}[!t]
\captionsetup{font=small, skip=5pt}
  \begin{center}
\includegraphics[scale = 0.4]{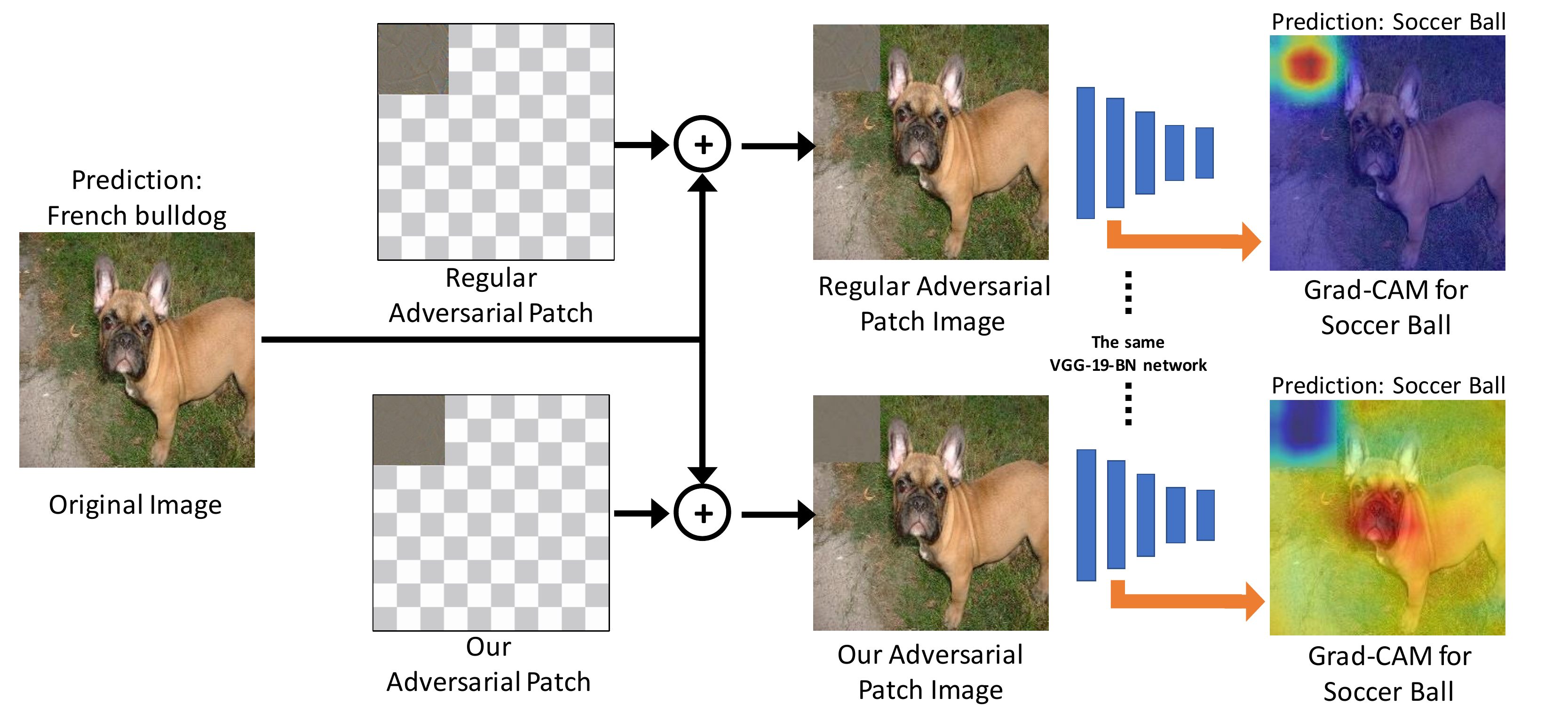}
  \caption{We show that Grad-CAM highlights the patch location in the image perturbed by regular targeted adversarial patches \cite{brown2017adversarial} (top row). Our modified attack algorithm goes beyond fooling the final prediction by also fooling the Grad-CAM visualization. Here, Grad-CAM is used to visualize the cause of the target category.}
\vspace{-0.3in}
\label{teaser}
  \end{center}
\end{figure*}

In this paper, we design adversarial attack algorithms that not only fool the network prediction but also fool the network interpretation. Our main goal is to utilize such attacks as a tool to investigate the reliability of network interpretation algorithms. Moreover, since our attacks fool the network interpretation, they can be seen as a potential vulnerability in the applications that utilize network interpretation to understand the cause of the prediction (e.g., in health-care applications \cite{ching2018opportunities}.)\\

\vspace{-0.1in}
\noindent {\bf Reliability of network interpretation:}
We are interested in studying the reliability of the interpretation in highlighting true cause of the prediction. To this end, we use the adversarial patch method \cite{brown2017adversarial} to design a {\em controlled} adversarial attack setting where the adversary changes the network prediction by manipulating only a small region of the image. Hence, we know that the cause of the wrong prediction should be inside the patch. We show that it is possible to optimize for an adversarial patch that attacks the prediction without being highlighted by the interpretation algorithm as the cause of the wrong prediction.

Grad-CAM \cite{selvaraju2016grad} is one of the most well-known network interpretation algorithms that performs well on sanity check among state-of-the-art interpretation algorithms recently studied in \cite{adebayo2018sanity}. Hence, we choose to study the correctness of Grad-CAM as a case study. Also, we show that our results even though tuned for Grad-CAM, can transfer directly to Occluding Patch \cite{zhou2014object} as another interpretation algorithm.

As an example, in Figure \ref{teaser}, the original image (left) is correctly classified as ``French Bulldog''. On the top row, a targeted adversarial patch has successfully changed the prediction to ``Soccer Ball''. Since the adversary is able to manipulate only the pixels inside the patch, it is expected that the interpretation algorithm (e.g, Grad-CAM) for ``Soccer Ball'' category should highlight some patch pixels as the cause of the wrong prediction. This is shown in the first row, top-right image. However, in the bottom row, our adversarial patch algorithm, not only changes the prediction to ``Soccer Ball'', but also does it in a way that Grad-CAM does not highlight the pixels inside the patch. We argue that since the adversary can change only the patch pixels, we know that the cause of the wrong prediction should be inside the patch. So, the observation that Grad-CAM does not highlight the patch pixels reveals that Grad-CAM is not reliably highlighting the source of prediction. Note that in this setting, the target category is not chosen by the model and is randomly chosen by the adversary from all possible wrong categories (i.e., uniformly from 999 categories of ImageNet). We believe this shows that the Grad-CAM algorithm is not {\em necessarily} showing the true cause of the prediction. \\
We optimize the patch by adding a new term in the optimization of adversarial patches that suppresses Grad-CAM activation at the location of the patch while still encouraging the wrong prediction (target category). We believe our algorithms can be used as a form of evaluation for future interpretation algorithms.\\

\vspace{-0.1in}
\noindent {\bf Practical implications:}
Our attack is more practical since we are manipulating only a patch and hence is closer to real world applications. As a practical example, some applications in health-care are not only interested in the prediction, but also understanding the cause of it (e.g., what region of a medical image of a patient causes diagnosis of cancer.) We believe our attacks can be generalized beyond object classification to empower an adversary to manipulate the reasoning about some medical diagnosis. Such an attack can cause serious issues in health-care, for instance by manipulating medical records to charge insurance companies \cite{Finlayson1287}.\\

\vspace{-.2in}
\noindent Our key contributions are summarized as follows:

{\bf (1)} We introduce a novel algorithm to construct adversarial patches which fool both the classifier and the interpretation of the resulting category.

{\bf (2)} With extensive experiments, we show that our method (a) generalizes from Grad-CAM to Occluding Patch \cite{zhou2014object}, another interpretation method, (b) generalizes to unseen images (universal), (c) is able to fool GAIN \cite{kunpeng2018gain}, a model specifically trained with supervision on interpretation, and (d) is able to make the interpretation uniform to hide any signature of the attack.

{\bf (3)} We use these attacks as a tool to assess the reliability of Grad-CAM, a popular network interpretation algorithm. This suggests that the community needs to develop more robust interpretation algorithms possibly using our tool as an evaluation method.

\section{Related work} \label{related_work_section}
\noindent {\bf Adversarial examples:}
Adversarial examples were discovered by Szegedy \etal \cite{intriguing-arxiv-2013} who showed that state-of-the-art machine learning classifiers can be fooled comprehensively by simple backpropagation algorithms. Goodfellow \etal \cite{explainingharnessing-arxiv-2014} improved this by Fast Gradient Sign Method (FGSM) that needs only one iteration of optimization. The possibility of extending these examples to the real world was shown in \cite{world-arxiv-2016,sharif2016accessorize} and \cite{athalye2017synthesizing} showed that adversarial examples could be robust to affine transformations as well. Madry \etal\cite{madry2017towards} proposed Projected Gradient Descent (PGD) which has been shown to be the best first-order adversary for fooling classifiers. Chen \etal\cite{chen2018robust} show that physical world adversarial examples can be created for object detection networks such as Faster R-CNN. Zaj{\k{a}}c \etal \cite{zajkac2018adversarial} also showed that instead of modifying the image, a frame can be placed around the image to fool classifiers. Su \etal \cite{su2019one} showed that modifying one pixel using Differential Evolution (DE) is sufficient to fool classifiers.
Although there have been many proposed defense algorithms, most of them have been overcome by making changes to the attack algorithm as shown in \cite{Carlini2017AdversarialEA,Athalye2018ObfuscatedGG}. Training robust networks is an important problem that can lead to better understanding of neural networks and also improve their generalization capabilities.

\noindent {\bf Adversarial patches:}
Adversarial patches \cite{brown2017adversarial,karmon2018lavan} were introduced as a more practical version of adversarial attacks where we restrict the spatial dimensions of the perturbation, but remove the imperceptibility constraint. These patches can be printed and `pasted' on top of an image to mislead classification networks. Recently, \cite{220580} showed that physical adversarial examples and adversarial patches can be created for object detection algorithms as well. We improve this by ensuring that the patches fool network interpretation tools that try to understand the reasoning for misclassification.

\noindent {\bf Interpretation of deep neural networks:}
As neural networks are getting closer towards deployment in real world applications, it is important that their results are interpretable.
Doshi-Velez \etal \cite{doshi2017accountability} discuss the legal and societal implications of explainable AI and suggest that although explainable systems might possibly be sub-optimal, it is a necessity that needs to be considered under design. This becomes extremely relevant when machine learning is used in biology, where it is essential to ensure the model's decision-making process is reliable and is not due to an artifact of the data (See Discussion in \cite{ching2018opportunities}). So it is important to make sure that the deep neural networks can be explained using robust and reliable interpretation algorithms which can ensure transparency in the network's explanation.
Researchers have proposed various algorithms in this direction. One of the earliest attempt \cite{simonyan2013deep} calculates the derivative of the network's outputs w.r.t the input to compute class specific saliency maps. Zhou \etal \cite{zhou2014object} calculates the change in the network output when a small portion of the image ($11\times11$ pixels) is covered by a random occluder. We call this \textbf{Occluding Patch}. CAM \cite{zhou2016learning} used weighted average map for each image based on their activations. The most popular one that we consider in this paper is called \textbf{Grad-CAM} \cite{selvaraju2016grad}, a gradient based method which provides visual explanations for any neural network architecture. Li \etal \cite{kunpeng2018gain} recently improved upon Grad-CAM using Guided attention mechanism with state-of-the-art results on PASCAL VOC 2012 segmentation task. Although the above methods have shown great improvement in explaining the network's decision, our work highlights that it is important to ensure that they are robust enough to adversaries as well.

\noindent \textbf{Attacking network interpretation:} Ghorbani \etal \cite{ghorbani2017interpretation} introduce adversarial perturbations that produce perceptively indistinguishable inputs that are assigned the \emph{same} predicted label, yet have very \emph{different} interpretations. However, in this setting, the adversarial image after perturbation can have image regions which correspond to stronger features for the same predicted label and as a result lead to different interpretations by dominating the prediction score. This is also noted in the discussion section in \cite{ghorbani2017interpretation}. To mitigate this, we design a {\em controlled} setting using adversarial patches where the adversary changes the network prediction by manipulating only a small region of the image. Here, we clearly know that the interpretation for the wrong prediction should be inside the patch. Heo \etal \cite{heo2019fooling} introduce a threat model wherein the adversary can modify the model parameters to fool the network interpretation. However, in a practical setting, the adversary might not always be able to modify the model parameters. Hence, we are interested in modifying only the pixels in a small image area without altering the model. Kindermans \etal \cite{kindermans2017reliability} showed how saliency methods are unreliable by adding a constant shift to input data and checking against different saliency methods. Adebayo \etal \cite{adebayo2018sanity} introduce sanity checks to evaluate existing saliency methods and show that some of them are independent of both the model and the data generating process. We believe our method can serve as an additional evaluation for future interpretation algorithms.

\section{Method} \label{method_section}

We propose algorithms to learn adversarial patches that when pasted on the input image, can change the interpretation of the model's prediction. We will focus on Grad-CAM \cite{selvaraju2016grad} in designing our algorithms and then, show that our results generalize to other interpretation algorithms as well. \\

\vspace{-0.1in}
\noindent {\bf Background on Grad-CAM visualization} %

Consider a deep network for image classification task, e.g., VGG, and an image $x_0$. We feed the image to the network and get the final output $y$  where $y^c$ is the logit or class-score for the $c$'th class. To interpret the network's decision for category $c$, we want to generate heatmap $G^c$ for a convolutional layer, e.g, {\em conv5}, which when up-sampled to the size of input image, highlights the regions of the image that have significant effect in producing higher values in $y^c$. We denote $A^k_{ij}$ as the activations of the $k$'th neuron at location $(i,j)$ of the chosen layer. Then, as in \cite{selvaraju2016grad}, we measure the effect of each feature of the convolutional layer at the final prediction by:
\vspace{-0.15in}
$$\alpha^c_k = \frac{1}{Z}\sum_{i}\sum_{j}{\frac{\partial y^c}{\partial A^k_{ij}}}$$
\vspace{-0.175in}

\noindent where $Z$ is a normalizer. Then we calculate the interpretation (heatmap) as the weighted sum of activations of the convolutional layer discarding the negative values:
\vspace{-0.075in}
$$G_{ij}^c = max(0, \sum_k\alpha^c_kA^k_{ij})$$
\vspace{-0.15in}

\noindent We then normalize the heatmap: $\displaystyle\hat{G}^c := \frac{G^c}{|G^c|_1}$ \\

 \vspace{0.05in}
\noindent {\bf Background on adversarial patches}

Consider an input image $x$ and a predefined constant binary mask $m$ that is $1$ on the location of the patch (top left corner in the experiments of Figure \ref{teaser}) and $0$ everywhere else. We want to find an adversarial patch $z$ that changes the output of the network to category $t$ when pasted on the image, so we solve:
\vspace{-0.05in}
$$ z = \argmin_{z} \ell_{ce}(x \odot (1-m)+z \odot m; t)$$
\vspace{-0.15in}

\noindent where $\ell_{ce}(.; t)$ is the cross entropy loss for the target category $t$
and $\odot$ is the element-wise product. Note that for simplicity of the notation, we assume $z$ has the same size as $x$, but only the patch location is involved in the optimization. This results in adversarial patches similar to \cite{brown2017adversarial}.

\subsection{Fooling interpretation with targeted patches}
\label{sec:target}
We now build upon the Grad-CAM method and adversarial patches explained in the preceding section to design our controlled setting that lets us study the reliability of network interpretation algorithms.
As shown in Figure \ref{teaser}, when an an image is attacked by an adversarial patch, Grad-CAM of the target category (wrong prediction) can be used to investigate the cause of the misclassification. It highlights the patch very strongly revealing the cause of the attack. This is expected as the adversary is restricted to perturbing only the patch area and the patch is the cause of the final misclassification towards target category.

In order to hide the adversarial patch in the interpretation of the final prediction, we add an additional term to our loss function while optimizing the patch such that the heatmap of the Grad-CAM interpretation at the patch location $m$ is suppressed. Hence, assuming the perturbed image \\ $\tilde x = x_0 \odot (1-m)+z \odot m$, we optimize:
\vspace{-0.05in}
\begin{equation}
\begin{split}
     \argmin_{z}\Big[\ell_{ce}(\tilde x; t) + \lambda \sum_{ij} \big(\hat{G}^t(\tilde x) \odot m\big)\Big ]
\end{split}
\label{eq1}
\end{equation}
\vspace{-0.125in}

\noindent where $t$ is the target category and $\lambda$ is the hyper-parameter to trade-off the effect of two loss terms. We choose the target label randomly across all classes excluding the original prediction similar to ``step rnd'' method in \cite{atscale-arxiv-2016}.

To optimize the above loss function, we use an iterative approach similar to projected gradient decent (PGD) algorithm \cite{madry2017towards}. We initialize $z$ randomly and iteratively update it by: $\displaystyle z^{n+1} = z^n - \eta Sign({\frac{\partial \ell}{\partial z}})$ with learning rate $\eta$.
At each iteration, we project $z$ to the feasible region by clipping it to the dynamic range of image values.

We argue that if this method succeeds in fooling the Grad-CAM to not highlight the adversarial patch location, it means the Grad-CAM algorithm is not showing the true cause of the attack since we know the attack is limited to the patch location only.

\subsection{Non-targeted patches}

A similar approach can be used to develop a non-targeted attack by maximizing the cross entropy loss of the correct category. This can be considered a weaker form of attack since the adversary has no control over the final category which is predicted after adding the patch. In this case, our optimization problem becomes:

\vspace{-.20in}
\begin{equation}
\begin{split}
     \argmin_{z}\Big[\max(0, M -\ell_{ce}(\tilde x; c)) \\
     + \enspace \lambda \sum_{ij} \big(\hat{G}^a(\tilde x) \odot m\big)\Big ]   %
\end{split}
\end{equation}
\vspace{-.15in}

\noindent where $c$ is the predicted category for the original image, $a=\argmax_k{y(k)}$ is the top prediction at every iteration, and $y(k)$ is the logit for category $k$. Since cross entropy loss is not upper-bounded, it can dominate the optimization, so we use contrastive loss \cite{hadsell2006dimensionality} to ignore cross entropy loss when the probability of $c$ is less than the chance level, thus $M=-log(p_0)$ where $p_0$ is the chance probability (e.g., 0.001 for ImageNet). Note that the second term is using the interpretation of the current top category $a$.

\subsection{Targeted regular adversarial examples}
\label{sec:non_patch_target}
We now consider regular adversarial examples (non-patch) \cite{intriguing-arxiv-2013} where the $\ell_\infty$ norm of the perturbation is restricted to a small {$\epsilon$}, (e.g. 8/255) which fools both the network prediction and the network interpretation. To this end, in Eq. \ref{eq1}, we expand  mask $m$ to cover the whole image and initialize $z$ from $x$. For completeness, we report the results of such attacks in our experiments, but as noted in the related work section, they do not necessarily show that the interpretation method is wrong.

\subsection{Universal targeted patches}
\label{universal_attention}
Universal attack is a much stronger form of attack wherein we train a patch that generalizes across images in fooling towards a particular category. Such an attack shows that it is possible to fool an unknown test image using a patch learned using the training data. This is a more practical form of attack, since the adversary needs to train the patch just once, which would be strong enough to fool multiple unseen test images. To do so, we optimize the summation of losses for all images in our training data using mini-batch gradient descent for:

\vspace{-.2in}
\begin{equation}
\begin{split}
     \argmin_{z}\sum_{n=1}^N \Big[\ell_{ce}(\tilde x_n; t) + \lambda \sum_{ij} \big(\hat{G}^t(\tilde x_n) \odot m\big)\Big ]
\end{split}
\end{equation}

\begin{table*}[!ht]
\captionsetup{font=small}
\centering
 \begin{tabular}{||c || c | c | c | c | c | c ||}
 \hline
 \multirow{2}{*}{Method} & Top-1 Acc(\%) & \multicolumn{2}{c|}{Non-Targeted} & \multicolumn{3}{c||}{Targeted} \\ [0.5ex]
 \cline{3-7}
 & & Acc (\%) & Energy Ratio (\%) & Acc (\%) & Target Acc (\%) & Energy Ratio(\%)\\
 \hline\hline
   Adversarial Patch \cite{brown2017adversarial} & 74.24 & 0.06 & 50.87 & 0.02 & 99.98 & 76.26   \\ %
 \hline
Our Patch & 74.24 & 0.05 & \textbf{2.61} & 2.95 &77.88 & \textbf{6.80} \\ %
 \hline
\end{tabular}
\newline
    \caption{Comparison of heatmap energy within the 8\% patch area for the adversarial patch \cite{brown2017adversarial} and our patch. We use an ImageNet pretrained VGG19-BN model on 50,000 images of the validation set of ImageNet dataset. Accuracy denotes the fraction of images that had the same final predicted label as the original image. Target Accuracy denotes the fraction of images where the final predicted label has changed to the randomly chosen target label.
    }
    \vspace{-.1in}
    \label{fig:comparison_patch_heatmap}
\end{table*}

\section{Experiments} \label{experiments_section}

We perform our experiments in two different benchmarks. We use VGG19 network with batch normalization, ResNet-34 and DenseNet-121 as standard network architectures. We use ImageNet \cite{deng2009imagenet} ILSVRC2012 for these experiments.

Then to evaluate our attack in a more challenging setting, we use GAIN{$_{ext}$} model from \cite{kunpeng2018gain} which is based on VGG19 (without batch normalization), but is specifically trained with supervision on the network attention to provide more accurate interpretation. We use PASCAL VOC-2012 dataset for these experiments since GAIN{$_{ext}$} uses semantic segmentation annotation and its pre-trained model is available only for this dataset.

\subsection{Evaluation}
We use standard classification accuracy to report the success rate of the attack and use the following metrics to measure the success of fooling interpretation:

{\bf (a) Energy Ratio:} We normalize the interpretation heatmap to sum to one for each image, and then calculate the ratio of the total energy of the interpretation at the patch location to that of the whole image. We call this metric ``Energy Ratio''. It will be $0$ if the patch is not highlighted at all and $1$ if the heatmap is completely concentrated inside the patch. In the case of a uniform heatmap, the energy ratio will be $8.2\%$ (the relative area of the patch).

{\bf (b) Histogram Intersection:} To compare two different interpretations, we calculate the Grad-CAM heatmap of the original image and the adversarial image, normalize each to sum to one per image, and calculate the histogram intersection between them.

{\bf (c) Localization:} We use the metric from the object localization challenge of ImageNet competition. Similar to the Grad-CAM paper, we draw a bounding box around values larger than a threshold (0.15 as used in \cite{selvaraju2016grad}), and evaluate object localization by comparing the boxes to the ground-truth bounding boxes.

We assume input images of size $224\times224$ and patches of size $64\times64$ which occupy almost $8.2\%$ of the image area. We place the patch on the top-left corner of the image for most experiments so that it does not overlap with the main objects of interest. We use PyTorch \cite{paszke2017automatic} along with NVIDIA Titan-X GPUs for all experiments.

\subsection{Targeted adversarial patches}
\label{exp_misleading_adv_patches}
For the adversarial patch experiments described in the method section, we use 50,000 images of the validation set of ImageNet ILSVRC2012 \cite{deng2009imagenet}. We perform 750 iterations of optimization with $\eta = 0.005$ and $\lambda=0.05$. We use the Energy Ratio metric for evaluation. The results in Table \ref{fig:comparison_patch_heatmap} show that our patch has significantly less energy in the patch area. However, this comes with some reduction in the targeted attack accuracy which can be attributed to the increased difficulty of the attack. Figure \ref{fig_patch_target1} shows the qualitative results.

\subsection{Non-targeted adversarial patches}

\label{sec_non_targeted_patch}
Here, we perform the non-targeted adversarial patch attack using 50,000 images of the validation set of ImageNet \cite{deng2009imagenet} ILSVRC2012. We perform 750 iterations with $\eta=0.005$ and $\lambda=0.001$. The results are shown in Table \ref{fig:comparison_patch_heatmap} and the qualitative results are included in the appendix.

\subsection{Different networks and patch locations}
\label{exp_generalization_across_arch}
Most of our experiments use models based on VGG19-BN and also place the patch on the top left-corner of the image. In this section, we evaluate our targeted adversarial patch attack algorithm on ResNet-34 \cite{He2016DeepRL} and DenseNet-121 \cite{Iandola2014DenseNetIE} by placing the patch on the top-right corner of the image. Similar to VGG experiments, both models are pretrained on ImageNet dataset. We use 5,000 random images from the ImageNet validation set to evaluate these attacks using the Energy Ratio metric which is presented in Table \ref{fig:comparison_patch_heatmap_resnet34}. Our patch fools the interpretation while reaching the target category in more than $90\%$ of the images. The qualitative results for these experiments are included in the appendix.

\begin{table}[!ht]
\captionsetup{font=small}
\centering
 \begin{tabular}{||c || c | c ||}
 \hline
 \multirow{2}{*}{Method} & \multicolumn{2}{c||}{Targeted} \\ [0.5ex]
 \cline{2-3}
 & \footnotesize{Target Acc (\%)} & \footnotesize{Energy Ratio (\%)}\\
 \hline\hline
Adv. Patch (R-34) & 100.0 & 61.9   \\ %
 \hline
 Our Patch (R-34) & 90.3 & \textbf{8.2} \\ %
 \hline
 \hline
 Adv. Patch (D-121) & 99.9 & 71.3   \\
 \hline
 Our Patch (D-121) & 93.6 & \textbf{5.3} \\
 \hline
\end{tabular}
\newline

    \caption{Comparison of Grad-CAM heatmap energy within the 8\% patch area (placed at the top-right corner) for different networks on 10\% randomly sampled ImageNet validation images. R-34 and D-121 refer to ResNet-34 and DenseNet-121 models respectively.
    }
    \vspace{-0.05in}
    \label{fig:comparison_patch_heatmap_resnet34}
\end{table}

\ignore{
We suspect that the reason our attack works
 (successful) the activations on the last conv layer have large receptive fields, This means that A single pixel could trigger activation in a far away location without activating the correct location.
 }

\subsection{Uniform heatmap patches}
\begin{figure*}[!h]
\captionsetup{font=small}
  \begin{center}
  \begin{tabular}{| c c c c c|}
\hline  \footnotesize Original & \footnotesize{Adv. Patch \cite{brown2017adversarial}} & \footnotesize{Adv. Patch - GCAM} & \footnotesize{Our Patch} & \footnotesize{Our Patch - GCAM}\\
 \hline
\vspace{-.08in}
&&&&\\
\vspace{-.05in}
\begin{sideways} \footnotesize Target: Bridegroom \end{sideways}
\includegraphics[width=.13\textwidth]{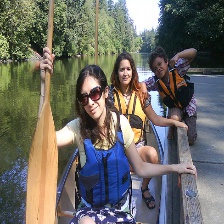}&
\includegraphics[width=.13\textwidth]{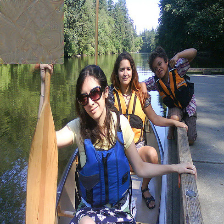}&
\includegraphics[width=.13\textwidth]{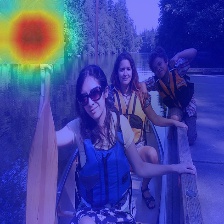}&
\includegraphics[width=.13\textwidth]{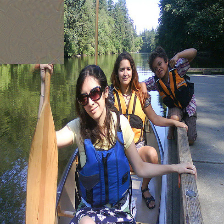}&
\includegraphics[width=.13\textwidth]{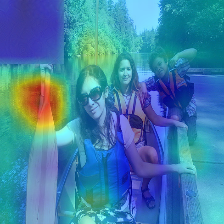}\\
\small Paddle & \small Bridegroom & \small Bridegroom & \small Bridegroom & \small Bridegroom \\

\vspace{-.05in}
\begin{sideways} \quad \small Target: Fig \end{sideways}
\includegraphics[width=.13\textwidth]{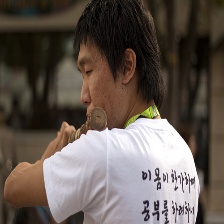}&
\includegraphics[width=.13\textwidth]{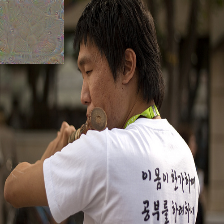}&
\includegraphics[width=.13\textwidth]{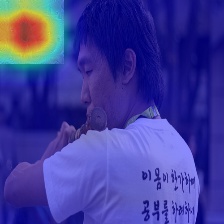}&
\includegraphics[width=.13\textwidth]{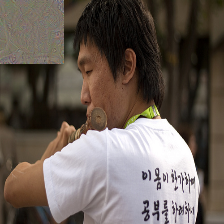}&
\includegraphics[width=.13\textwidth]{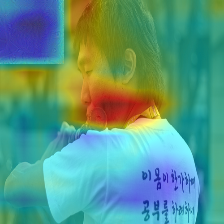}\\
\small Whistle & \small Fig & \small Fig & \small Fig & \small Fig \\
\hline
\end{tabular}
\vspace{.02in}
  \caption{\textbf{Targeted patch attack:} We use an ImageNet pretrained VGG 19 BN network to compare the Grad-CAM visualization results for a random target category using our method vs Adv. Patch \cite{brown2017adversarial}. The predicted label is written under each image. Note that the patch is not highlighted in the last column.}
\vspace{-.15in}
\label{fig_patch_target1}
  \end{center}
\end{figure*}

\begin{figure*}[!h]
\captionsetup{font=small}
  \begin{center}
  \begin{tabular}{| c c c c c|}
\hline  \footnotesize Original & \footnotesize{Adv. Patch \cite{brown2017adversarial}} & \footnotesize{Adv. Patch - GCAM} & \footnotesize{Our Patch} & \footnotesize{Our Patch - GCAM}\\
 \hline
\vspace{-.08in}
&&&&\\
\vspace{-.05in}
\begin{sideways} \small Target: Guillotine \end{sideways}
\includegraphics[width=.13\textwidth]{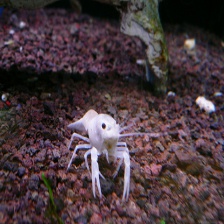}&
\includegraphics[width=.13\textwidth]{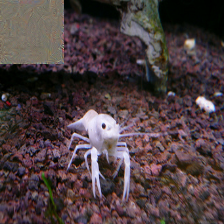}&
\includegraphics[width=.13\textwidth]{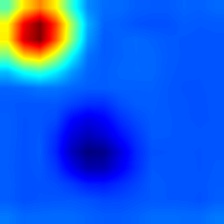}&
\includegraphics[width=.13\textwidth]{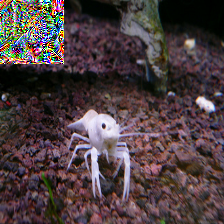}&
\includegraphics[width=.13\textwidth]{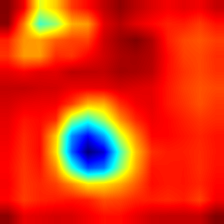}\\
\small Fiddler Crab & \small Guillotine & \small Guillotine & \small Guillotine & \small Guillotine \\

\vspace{-.05in}
\begin{sideways} \enspace \small Target: Velvet \end{sideways}
\includegraphics[width=.13\textwidth]{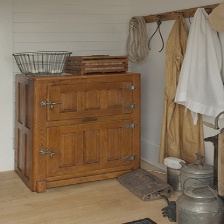}&
\includegraphics[width=.13\textwidth]{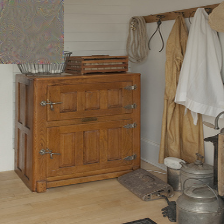}&
\includegraphics[width=.13\textwidth]{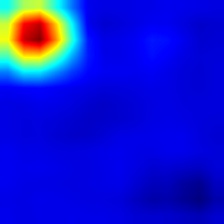}&
\includegraphics[width=.13\textwidth]{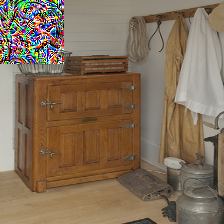}&
\includegraphics[width=.13\textwidth]{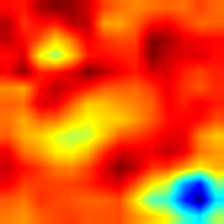}\\
\small Wardrobe & \small Velvet & \small Velvet & \small Velvet & \small Velvet \\
\hline
\end{tabular}
\vspace{.02in}
  \caption{\textbf{Uniform patch attack:} Here, we paste our adversarial patch on the top-left corner and encourage the Grad-CAM heatmap for the target category to highlight the top-right corner. This shows that our algorithm can also be modified to hide our patch in the Grad-CAM visualization. The predicted label is written under each image. Note that the patch is not identifiable in the last column.}
 \vspace{-.25in}
\label{fig_patch_target_uniform}
  \end{center}
\end{figure*}

One may argue that our attacks may not be effective in practice to fool the manual investigation of the network output since the lower (blue) heatmap of the Grad-CAM can still be considered as a distinguishable signature (see Figure \ref{fig_patch_target1}).
We mitigate this concern by optimizing the patch to encourage higher values of Grad-CAM outside the patch area (top-right corner instead of the patch area which is at the top-left corner). Our results in Table \ref{fig:comparison_patch_uniform_heatmap_GAIN} and Figure \ref{fig_patch_target_uniform} show that our attack can still fool the interpretation by generating a more uniform pattern for the heatmap. We perform 1,000 iterations with $\eta = 0.007$ and $\lambda=0.75$.

\begin{table}[!ht]
\captionsetup{font=small}
\centering
 \begin{tabular}{||c || c | c | c||}
 \hline
 \multirow{2}{*}{Method} & \multirow{2}{*}{\footnotesize{Target Acc (\%)}} & \multicolumn{2}{c||}{\footnotesize{Energy Ratio (\%)}}\\
 \cline{3-4}
 & & \footnotesize Top-Left & \footnotesize Top-Right\\
 \hline\hline
 \footnotesize Adv. Patch \cite{brown2017adversarial} & 100 & 76.96 & 1.65\\ %
 \hline
 \footnotesize Our Patch (Top-Left) & 83.5 & \textbf{14.99} & \textbf{7.57}\\
 \hline
\end{tabular}
\newline
    \caption{Comparison of heatmap energy for the uniform patches. \ignore{Here, we place our adversarial patch on the top-left corner and encourage the Grad-CAM heatmap to highlight a different location (top-right). This encourages our patch to be not easily identifiable due to the very low heatmap values.} We report the energy at both the top-left and top-right corners of the heatmap.
    }
    \label{fig:comparison_patch_uniform_heatmap_GAIN}
\vspace{-0.15in}
\end{table}

\subsection{Targeted regular adversarial examples}
\label{exp_misleading_adv_examples}
For the regular adversarial examples (non-patch) described in Section \ref{sec:non_patch_target} that fool both the network prediction and interpretation, we perform 150 iterations with $\epsilon=8/255$, $\eta=0.001$, and $\lambda=0.05$. Since the attack is not constrained to a patch location, the Energy Ratio metric is no longer applicable in this case. We use Localization Error and Histogram Intersection as the evaluation metrics in Table \ref{table:PGD}. We compare with PGD attack \cite{madry2017towards} as a baseline. The corresponding qualitative results are included in the appendix. Note that in this case, we run Grad-CAM for the original predicted category.

\begin{table}[!ht]
\captionsetup{font=small}
\centering
\begin{tabular}{||c || c | c ||}
 \hline
 Image & Loc. Error(\%) & Histogram\\
 \hline
 \hline
 Original  & 66.68 & 1.0 \\
 \hline
 PGD Adv. & 67.74 & 0.77 \\
 \hline
 Grad-CAM Adv. & \textbf{76.02} & \textbf{0.64} \\
 \hline
\end{tabular}
\caption{Evaluation results for adversarial examples generated using our method and PGD \cite{madry2017towards} on 10\% randomly sampled ImageNet validation images. Note that for histogram intersection, lower is better while for localization error, higher is better. }
\label{table:PGD}
\vspace{-0.1in}
\end{table}

\subsection{Targeted patch on guided attention models}
\begin{figure*}[!ht]
\captionsetup{font=small}
  \begin{center}
  \begin{tabular}{| c c c c c|}
\hline  \footnotesize Original & \footnotesize{Adv. Patch \cite{brown2017adversarial}} & \footnotesize{Adv. Patch - GCAM} & \footnotesize{Our Patch} & \footnotesize{Our Patch - GCAM}\\
\hline
\vspace{-.08in}
&&&&\\
\vspace{-.05in}
\begin{sideways} \quad \small Target: Sofa \end{sideways}
\includegraphics[width=.13\textwidth]{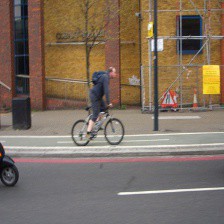}&
\includegraphics[width=.13\textwidth]{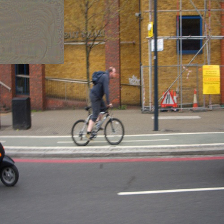}&
\includegraphics[width=.13\textwidth]{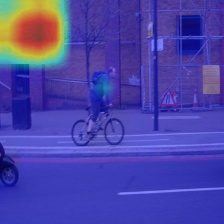}&
\includegraphics[width=.13\textwidth]{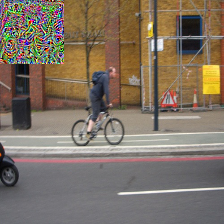}&
\includegraphics[width=.13\textwidth]{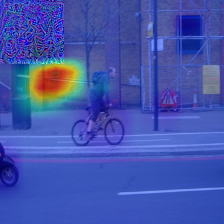}\\
\small Person & \small Sofa & \small Sofa & \small Sofa & \small Sofa \\

\vspace{-.05in}
\begin{sideways} \quad \small Target: Chair \end{sideways}
\includegraphics[width=.13\textwidth]{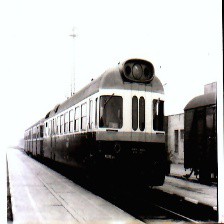}&
\includegraphics[width=.13\textwidth]{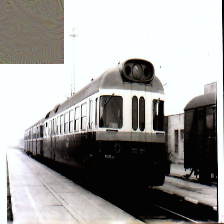}&
\includegraphics[width=.13\textwidth]{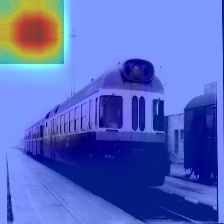}&
\includegraphics[width=.13\textwidth]{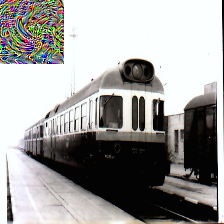}&
\includegraphics[width=.13\textwidth]{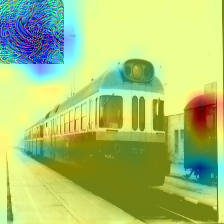}\\
\small Train & \small Chair & \small Chair & \small Chair & \small Chair \\

\hline
\end{tabular}
\vspace{.02in}
  \caption{\textbf{Targeted attack for guided attention models:} We use GAIN{$_{ext}$} \cite{kunpeng2018gain} VGG19 model on VOC dataset to compare Grad-CAM visualization results for the least likely target category using our method vs Adv. Patch \cite{brown2017adversarial}. The predicted label is written under each image. GAIN{$_{ext}$} is particularly designed to produce better Grad-CAM visualizations using direct supervision on the Grad-CAM output. }
\vspace{-.18in}
\label{fig_patch_target1_GAIN}
  \end{center}
\end{figure*}

To challenge our attack algorithms, we use the GAIN{${_{ext}}$} model \cite{kunpeng2018gain} which is based on VGG19 and is supervised using semantic segmentation annotation to produce better Grad-CAM results. The model is pre-trained on the training set of PASCAL VOC-2012, and we use the test set for optimizing the attack. Since each image in the VOC dataset can contain more than one category, we use the least likely predicted category as the target category. We perform 750 iterations with $\eta=0.1$ and $\lambda= 10^{-5}$. The qualitative results are shown in Figure \ref{fig_patch_target1_GAIN} and the quantitative ones in Table \ref{fig:comparison_patch_heatmap_GAIN}. Interestingly, our attack can fool this model even though it is trained to provide better Grad-CAM results.

\begin{table}[!ht]
\captionsetup{font=small}
\centering
 \begin{tabular}{||c || c | c ||}
 \hline
 Method & Target Acc (\%) & Energy Ratio (\%)\\
 \hline\hline
 Adv. Patch \cite{brown2017adversarial} & 94.34 & 37.90 \\ %
 \hline
 Our Patch & 94.70 & \textbf{3.2} \\ %
 \hline
\end{tabular}
\newline
    \caption{Targeted adversarial patch attack on GAIN{$_{ext}$} model \cite{kunpeng2018gain}
    }
    \label{fig:comparison_patch_heatmap_GAIN}
\end{table}

\begin{figure*}[!h]
\captionsetup{font=small}
  \begin{center}
  \begin{tabular}{| c c c c c|}
\hline  \footnotesize Original & \footnotesize{Adv. Patch \cite{brown2017adversarial}} & \footnotesize{Adv. Patch} & \footnotesize Our Patch & \footnotesize Our Patch \\
& \footnotesize GCAM & \footnotesize Occluding Patch & \footnotesize GCAM & \footnotesize Occluding Patch \\
\hline
\vspace{-.08in}
&&&&\\
\vspace{-.05in}
\begin{sideways} \scriptsize Target: Dining Table \end{sideways}
\includegraphics[width=.13\textwidth]{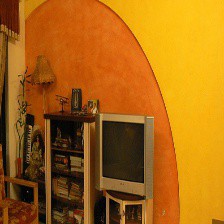}&
\includegraphics[width=.13\textwidth]{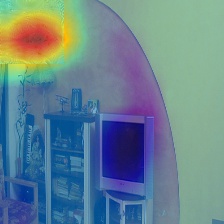}&
\includegraphics[width=.13\textwidth]{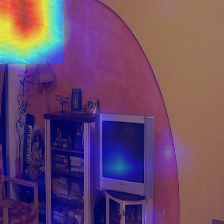}&
\includegraphics[width=.13\textwidth]{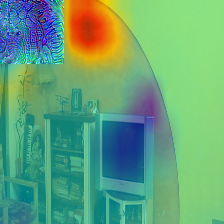}&
\includegraphics[width=.13\textwidth]{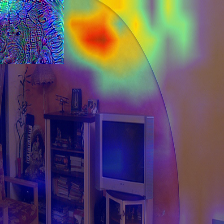}\\
\small TV / Monitor & \small Dining Table & \small Dining Table & \small Dining Table & \small Dining Table \\

\vspace{-.05in}
\begin{sideways}  \scriptsize Target: Potted Plant \end{sideways}
\includegraphics[width=.13\textwidth]{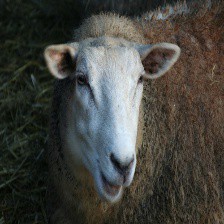}&
\includegraphics[width=.13\textwidth]{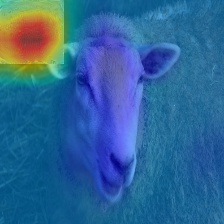}&
\includegraphics[width=.13\textwidth]{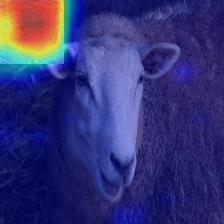}&
\includegraphics[width=.13\textwidth]{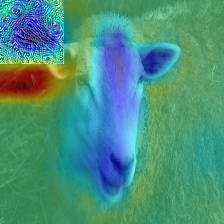}&
\includegraphics[width=.13\textwidth]{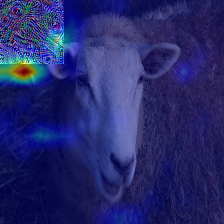}\\
\small Sheep & \small Potted Plant & \small Potted Plant & \small Potted Plant & \small Potted Plant \\
\hline
\end{tabular}
\vspace{.05in}
  \caption{\textbf{Generalization beyond Grad-CAM:} Transfer of Grad-CAM visualization attack to Occluding Patch visualization. Here, we use targeted patch attacks (least likely target category) using our method vs Adv. Patch \cite{brown2017adversarial} on the GAIN{$_{ext}$} \cite{kunpeng2018gain} network for VOC dataset. The predicted label is written under each image.  Grad-CAM and Occluding Patch visualizations are always computed for the target category. Note that the patch is hidden in both visualizations in columns 4 and 5.}
\label{fig_transfer}
  \end{center}
\end{figure*}

\subsection{Generalization beyond Grad-CAM}
We show that our patches learned using Grad-CAM are also hidden in the visualizations generated by Occluding Patch \cite{zhou2014object} method, which is a different interpretation algorithm. In occluding patch method, we visualize the change in the final score of the model by sliding a small black box on the image. Larger decrease in the score indicates that the regions are more important and hence they contribute more to the heatmap. The results of fooling GAIN{$_{ext}$} model are shown in Table \ref{table:comparison_patch_transfer_heatmap_GAIN} and Figure \ref{fig_transfer}.

\begin{table}[!ht]
\captionsetup{font=small}
\centering
 \begin{tabular}{||c || c | c ||}
 \hline
 Method & \small{Targeted Attack Energy Ratio (\%)}\\ [0.5ex]
 \hline\hline
   Adversarial Patch \cite{brown2017adversarial} & 80.44  \\ %

 \hline
 Our Patch & \textbf{31.59}  \\ %
 \hline
\end{tabular}
\newline
    \caption{Results showing transfer of our patch trained for Grad-CAM and evaluated on Occluding Patch \cite{zhou2014object} visualization using the GAIN{$_{ext}$} model for VOC dataset. \ignore{heatmap energy within the 8\% patch area for the adversarial patch \cite{brown2017adversarial} and our patch trained on the GAIN{$_{ext}$} \cite{kunpeng2018gain} for VOC dataset. Note that we still use Grad-CAM in training and evaluate on Occluding Patch. This shows our attack generalizes from Grad-CAM to occluding patch.}
    }
    \label{table:comparison_patch_transfer_heatmap_GAIN}
\end{table}

\begin{figure*}[h!]
\captionsetup{font=small}
  \begin{center}
  \begin{tabular}{| c c c c c|}
\hline  \footnotesize Original & \footnotesize{Adv. Patch \cite{brown2017adversarial}} & \footnotesize{Adv. Patch - GCAM} & \footnotesize Our Patch & \footnotesize Our Patch - GCAM \\
  \hline
\vspace{-.08in}
&&&&\\
\vspace{-.05in}
\begin{sideways} \footnotesize Target: Aeroplane \end{sideways}
\includegraphics[width=.13\textwidth]{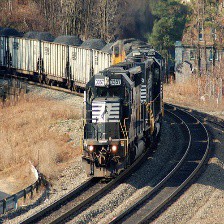}&
\includegraphics[width=.13\textwidth]{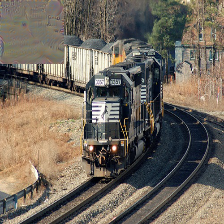}&
\includegraphics[width=.13\textwidth]{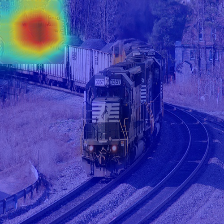}&
\includegraphics[width=.13\textwidth]{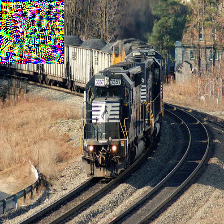}&
\includegraphics[width=.13\textwidth]{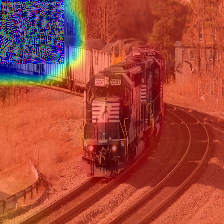}\\
\small Train  & \small Aeroplane & \small Aeroplane & \small Aeroplane & \small Aeroplane\\
\hline

\end{tabular}
\vspace{.05in}
  \caption{\textbf{Universal targeted patch:} Grad-CAM visualization results comparing our method vs Adv. Patch \cite{brown2017adversarial}. The top-1 predicted label is written under each image and Grad-CAM is always computed for the target category. The target category chosen was ``Aeroplane''. Additional results are included in the appendix.}
\label{fig_patch_universal}
\end{center}
\end{figure*}

\begin{table*}[h!]
\captionsetup{font=small}
\centering
 \begin{tabular}{||c | c || c | c | c | c | c | c | c | c | c | c | c | c | c | c | c | c | c | c | c | c ||}
 \hline
 \multicolumn{2}{|c|}{\scriptsize Methods} & \tiny aero & \tiny bike & \tiny bird & \tiny boat & \tiny bottle & \tiny bus & \tiny car & \tiny cat & \tiny chair & \tiny cow & \tiny table & \tiny dog & \tiny horse & \tiny mbike & \tiny person & \tiny plant & \tiny sheep & \tiny sofa & \tiny train & \tiny tv \\
 \hline\hline
 \multirow{2}{*}{\begin{sideways}\centering \scriptsize Energy \end{sideways}} &
 \tiny Reg. Patch & \tiny 86.3 & \tiny 64.8 & \tiny 90.3 & \tiny 47.7 & \tiny 92.8 & \tiny \textbf{0.0} & \tiny 78.3 & \tiny 48.4 & \tiny 84.5 & \tiny 94.4 & \tiny 78.2 & \tiny 90.9 & \tiny 16.9 & \tiny 85.0 & \tiny 78.61 & \tiny 1.1 & \tiny 86.9 & \tiny 85.4 & \tiny 83.0 & \tiny 98.2 \\
 \cline{2-22}
 & \tiny Our Patch & \tiny \textbf{0.0} & \tiny \textbf{0.8} & \tiny \textbf{0.6} & \tiny \textbf{0.3} & \tiny \textbf{0.0} & \tiny 0.2 & \tiny \textbf{1.4} & \tiny \textbf{0.6} & \tiny \textbf{0.6} & \tiny \textbf{2.4} & \tiny \textbf{3.7} & \tiny \textbf{0.0} & \tiny \textbf{0.0} & \tiny \textbf{1.2} & \tiny \textbf{0.8} & \tiny \textbf{0.0} & \tiny \textbf{2.7} & \tiny \textbf{2.6} & \tiny \textbf{0.1} & \tiny \textbf{0.0} \\
 \hline

  \multirow{2}{*}
  {\begin{sideways} \scriptsize Target \end{sideways}
  \begin{sideways} \scriptsize Acc \end{sideways}} &
 \tiny Reg. Patch & \tiny 99.4 & \tiny 97.0 & \tiny 100 & \tiny 98.8 & \tiny 100 & \tiny 92.6 & \tiny 93.2 & \tiny 99.8 & \tiny 99.3 & \tiny 100 & \tiny 99.0 & \tiny 100 & \tiny 99.9 & \tiny 99.2 & \tiny 99.8 & \tiny 98.7 & \tiny 99.8 & \tiny 99.6 & \tiny 99.5 & \tiny 100 \\
 \cline{2-22}
 & \tiny Our Patch & \tiny 94.5 & \tiny 97.4 & \tiny 99.3 & \tiny 84.5 & \tiny 94.3 & \tiny 99.9 & \tiny 99.7 & \tiny 99.6 & \tiny 98.7 & \tiny 34.3 & \tiny 87.3 & \tiny 94.7 & \tiny 98.3 & \tiny 99.4 & \tiny 99.8 & \tiny 99.2 & \tiny 99.8 & \tiny 93.8 & \tiny 99.2 & \tiny 99.0 \\
 \hline
\end{tabular}
\newline
    \caption{Results for the universal targeted patch attack using the GAIN{$_{ext}$} \cite{kunpeng2018gain} model on PASCAL VOC-2012 dataset using regular adversarial patch \cite{brown2017adversarial} and our adversarial patch. We learn universal patches for each of the 20 classes as the target category.}
    \label{fig:comparison_patch_univ_heatmap_GAIN}
\vspace{-.1in}
\end{table*}

\subsection{Universal targeted patches}
\label{sec:universal}
To show that the patch can generalize across images, we learn a universal patch for a given category using the training data and evaluate it on the test data. We use GAIN{$_{ext}$} model along with $\eta=0.05$ and $\lambda = 0.09$. The results are shown in Figure \ref{fig_patch_universal} and Table \ref{fig:comparison_patch_univ_heatmap_GAIN}. We learn 20 different patches for each class of PASCAL VOC-2012 as the target category. We observe high fooling rates for both our method and regular adversarial patch, but our method has considerably low energy focused inside the patch area.

\section{Conclusion} \label{conclusion_section}
We introduce adversarial patches (small area, \texttildelow 8\%, with unrestricted perturbations) which fool both the classifier and the interpretation of the resulting category. Since we know that the patch is the true cause of the wrong prediction, a reliable interpretation algorithm should definitely highlight the patch region. We successfully design an adversarial patch that does not get highlighted in the interpretation and hence show that popular interpretation algorithms are not highlighting the true cause of the prediction.  %
Moreover, we show that our attack works in various settings: (1) generalizes from Grad-CAM to Occluded Patch \cite{zhou2014object}, another interpretation method, (2) generalizes to unseen images (universal), (3) is able to fool GAIN \cite{kunpeng2018gain}, a model specifically trained with supervision on interpretation and (4) is able to make the interpretation uniform to hide the signature of the attack. Our work suggests that the community needs to develop more robust interpretation algorithms. \\

\vspace{-.15in}
\noindent {\bf Acknowledgement:} This work was performed under the following financial assistance award: 60NANB18D279 from U.S. Department of Commerce, National Institute of Standards and Technology, funding from SAP SE, and also NSF grant 1845216.

\clearpage
{\small
\bibliographystyle{ieee_fullname}
\bibliography{egbib}
}
\clearpage
\section{Appendix}

\begin{figure}[!h]
\setlength{\linewidth}{\textwidth}
\setlength{\hsize}{\textwidth}
  \begin{center}
  \begin{tabular}{| c c c c c|}
  \hline
  \footnotesize Original & \footnotesize Adv. Patch & \footnotesize Adv. Patch - GCAM & \footnotesize Our Patch & \footnotesize Our Patch - GCAM\\
\hline
&&&&\\
\begin{sideways} \enspace \footnotesize Target: Tray \end{sideways}
\includegraphics[width=.135\textwidth]{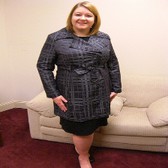}&
\includegraphics[width=.135\textwidth]{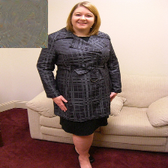}&
\includegraphics[width=.135\textwidth]{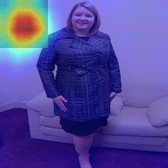}&
\includegraphics[width=.135\textwidth]{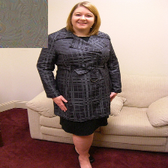}&
\includegraphics[width=.135\textwidth]{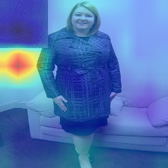}\\
\enspace Cardigan  & Tray & Tray & Tray & Tray \\
&&&&\\
\begin{sideways}\parbox{2cm}{{\scriptsize Target:}{\scriptsize Flat-coated Retriever}}\end{sideways}
\includegraphics[width=.135\textwidth]{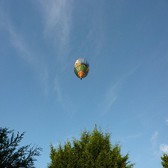}&
\includegraphics[width=.135\textwidth]{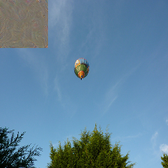}&
\includegraphics[width=.135\textwidth]{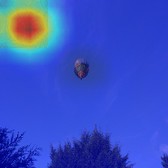}&
\includegraphics[width=.135\textwidth]{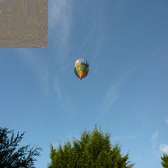}&
\includegraphics[width=.135\textwidth]{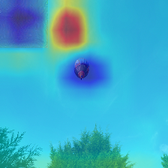}\\
\enspace \quad Balloon & Flat-coated  &Flat-coated  & Flat-coated  & Flat-coated \\
 & Retriever & Retriever &  Retriever &  Retriever\\
 &&&&\\
\begin{sideways} \enspace \quad \footnotesize Target: Tick \end{sideways}
\includegraphics[width=.135\textwidth]{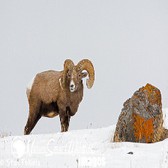}&
\includegraphics[width=.135\textwidth]{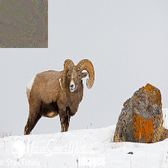}&
\includegraphics[width=.135\textwidth]{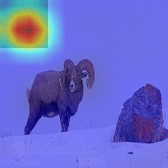}&
\includegraphics[width=.135\textwidth]{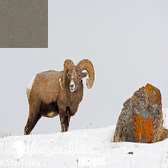}&
\includegraphics[width=.135\textwidth]{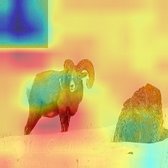}\\
Ram & Tick & Tick & Tick & Tick\\
&&&&\\
\begin{sideways} \enspace \quad \footnotesize Target: Stole \end{sideways}
\includegraphics[width=.135\textwidth]{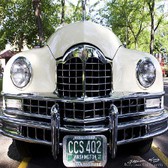}&
\includegraphics[width=.135\textwidth]{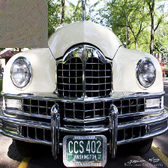}&
\includegraphics[width=.135\textwidth]{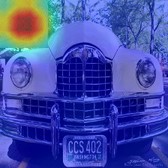}&
\includegraphics[width=.135\textwidth]{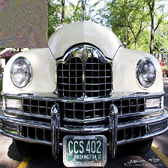}&
\includegraphics[width=.135\textwidth]{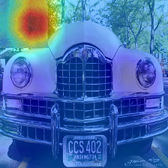}\\
\quad Radiator grille & Stole & Stole & Stole & Stole \\
\hline
\end{tabular}
  \caption{{\bf Targeted Patch Attack:} Additional results for targeted patch attack for ImageNet pretrained VGG19-BN network similar to Figure \ref{fig_patch_target1}. Images come from ImageNet validation set. The last row shows a failure case where our patch is not completely hidden in the interpretation.}
\label{fig_patch_target2}
  \end{center}
\end{figure}

\begin{figure*}[!ht]
  \begin{center}
  \begin{tabular}{| c c c c c|}
    \hline
\footnotesize Original & \footnotesize Adv. Patch & \footnotesize Adv. Patch - GCAM & \footnotesize Our Patch & \footnotesize Our Patch - GCAM\\
  \hline
&&&&\\
\includegraphics[width=.16\textwidth]{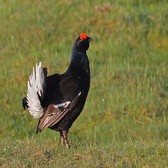}&
\includegraphics[width=.16\textwidth]{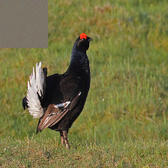}&
\includegraphics[width=.16\textwidth]{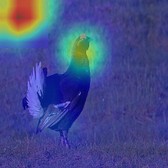}&
\includegraphics[width=.16\textwidth]{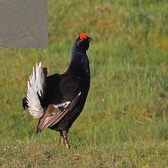}&
\includegraphics[width=.16\textwidth]{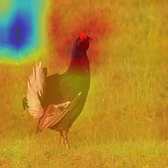}\\
Black grouse & Partridge & Partridge & Baseball & Baseball \\
 &&&&\\
\includegraphics[width=.16\textwidth]{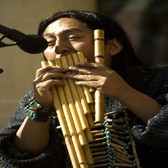}&
\includegraphics[width=.16\textwidth]{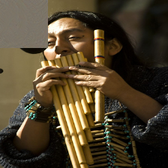}&
\includegraphics[width=.16\textwidth]{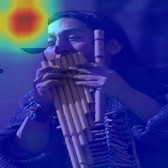}&
\includegraphics[width=.16\textwidth]{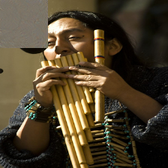}&
\includegraphics[width=.16\textwidth]{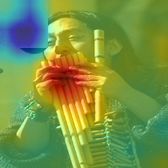}\\
Panpipe & Bath tissue & Bath tissue & Hornbill & Hornbill \\
&&&&\\
\includegraphics[width=.16\textwidth]{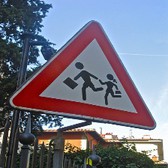}&
\includegraphics[width=.16\textwidth]{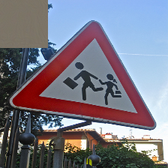}&
\includegraphics[width=.16\textwidth]{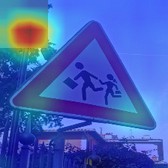}&
\includegraphics[width=.16\textwidth]{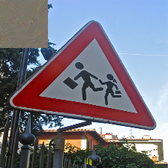}&
\includegraphics[width=.16\textwidth]{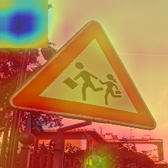}\\
Street sign & Lampshade & Lampshade & Patio & Patio \\
&&&&\\
\includegraphics[width=.16\textwidth]{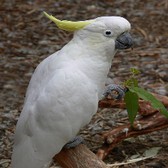}&
\includegraphics[width=.16\textwidth]{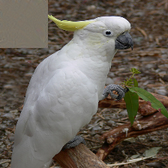}&
\includegraphics[width=.16\textwidth]{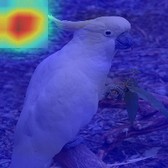}&
\includegraphics[width=.16\textwidth]{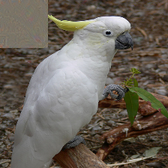}&
\includegraphics[width=.16\textwidth]{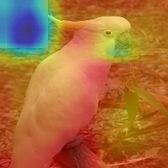}\\
Cockatoo  & Lycaenid & Lycaenid & Stinkhorn & Stinkhorn \\
\hline
\end{tabular}
\vspace{.05in}
  \caption{{\bf Non-targeted Patch Attack:} Comparison of Grad-CAM results for non-targeted patch attacks using our method vs regular adversarial patch. The description is in  Section \ref{sec_non_targeted_patch} and the quantitative results are in Table \ref{fig:comparison_patch_heatmap}. We use ImageNet pre-trained VGG19-BN. The predicted label is written under each image, the non-targeted attack was successful for all images, and Grad-CAM is always computed for the predicted category. Images come from ImageNet validation set.}
\label{fig_patch_nontarget1}
  \end{center}
\end{figure*}

\begin{figure*}[!ht]
  \begin{center}
  \begin{tabular}{| c c c c c|}
    \hline
\footnotesize Original & \footnotesize Adv. Patch & \footnotesize Adv. Patch - GCAM & \footnotesize Our Patch & \footnotesize Our Patch - GCAM\\
  \hline
&&&&\\
\begin{sideways} \enspace \footnotesize Target: White Shark \end{sideways}
\includegraphics[width=.16\textwidth]{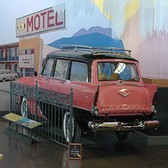}&
\includegraphics[width=.16\textwidth]{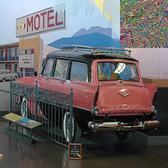}&
\includegraphics[width=.16\textwidth]{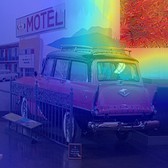}&
\includegraphics[width=.16\textwidth]{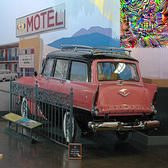}&
\includegraphics[width=.16\textwidth]{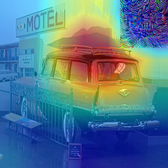}\\
Beach Wagon & White Shark & White Shark & White Shark & White Shark \\
&&&&\\
\begin{sideways} \footnotesize Target: Standard Poodle \end{sideways}
\includegraphics[width=.16\textwidth]{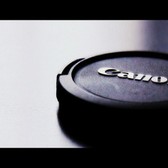}&
\includegraphics[width=.16\textwidth]{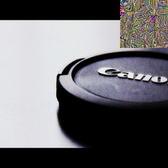}&
\includegraphics[width=.16\textwidth]{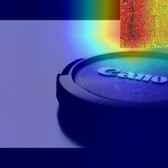}&
\includegraphics[width=.16\textwidth]{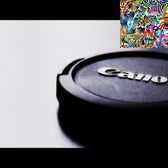}&
\includegraphics[width=.16\textwidth]{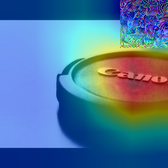}\\
Lens Cap & Standard Poodle & Standard Poodle & Standard Poodle & Standard Poodle \\
&&&&\\
\begin{sideways} \enspace \enspace \enspace \footnotesize Target: Thimble \end{sideways}
\includegraphics[width=.16\textwidth]{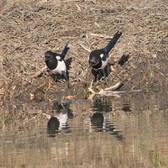}&
\includegraphics[width=.16\textwidth]{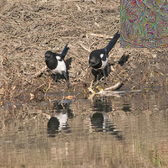}&
\includegraphics[width=.16\textwidth]{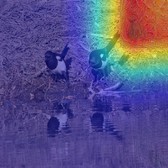}&
\includegraphics[width=.16\textwidth]{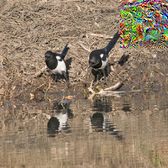}&
\includegraphics[width=.16\textwidth]{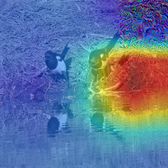}\\
Magpie & Thimble & Thimble & Thimble & Thimble \\
&&&&\\
\begin{sideways} \footnotesize Target: Wooden Spoon \end{sideways}
\includegraphics[width=.16\textwidth]{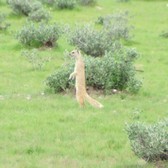}&
\includegraphics[width=.16\textwidth]{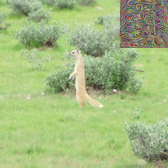}&
\includegraphics[width=.16\textwidth]{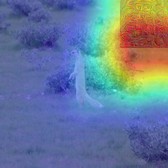}&
\includegraphics[width=.16\textwidth]{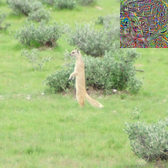}&
\includegraphics[width=.16\textwidth]{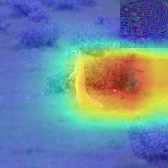}\\
Mongoose & Wooden Spoon & Wooden Spoon & Wooden Spoon & Wooden Spoon \\
  \hline
\end{tabular}
  \caption{{\bf Different networks and patch locations:} Comparison of Grad-CAM visualization results for our targeted patch attack vs regular adversarial patch. It uses ImageNet pretrained {\bf ResNet-34} network with the patch on the top right corner. The description is in  Section \ref{exp_generalization_across_arch} and the quantitative results are in Table  \ref{fig:comparison_patch_heatmap_resnet34}. The predicted label is written under each image, the targeted attack was successful for all images in this figure, and Grad-CAM is always computed for the target category. Images are from ImageNet validation set.}
\label{fig_patch_nontarget1}
  \end{center}
\end{figure*}

\begin{figure*}[!ht]
  \begin{center}
  \begin{tabular}{| c c c c c|}
    \hline
\footnotesize Original & \footnotesize Adv. Patch & \footnotesize Adv. Patch - GCAM & \footnotesize Our Patch & \footnotesize Our Patch - GCAM\\
  \hline
&&&&\\
\begin{sideways} \enspace \footnotesize Target: Corkscrew \end{sideways}
\includegraphics[width=.16\textwidth]{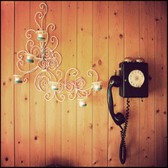}&
\includegraphics[width=.16\textwidth]{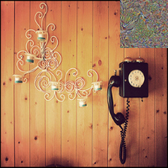}&
\includegraphics[width=.16\textwidth]{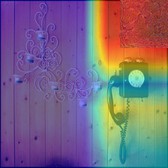}&
\includegraphics[width=.16\textwidth]{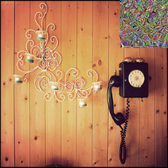}&
\includegraphics[width=.16\textwidth]{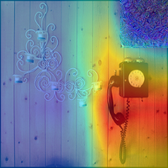}\\
Dial Phone & Corkscrew & Corkscrew & Corkscrew & Corkscrew \\
&&&&\\
\begin{sideways} \enspace \enspace \footnotesize Target: Toilet Seat \end{sideways}
\includegraphics[width=.16\textwidth]{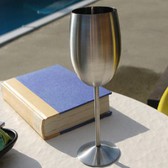}&
\includegraphics[width=.16\textwidth]{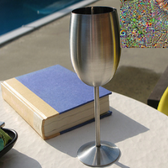}&
\includegraphics[width=.16\textwidth]{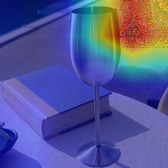}&
\includegraphics[width=.16\textwidth]{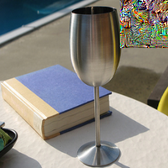}&
\includegraphics[width=.16\textwidth]{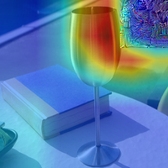}\\
Goblet & Toilet Seat & Toilet Seat & Toilet Seat & Toilet Seat \\
&&&&\\
\begin{sideways} \enspace \enspace \footnotesize Target: Doberman \end{sideways}
\includegraphics[width=.16\textwidth]{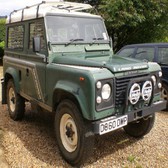}&
\includegraphics[width=.16\textwidth]{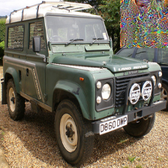}&
\includegraphics[width=.16\textwidth]{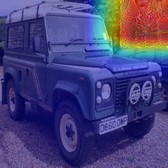}&
\includegraphics[width=.16\textwidth]{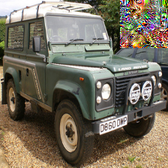}&
\includegraphics[width=.16\textwidth]{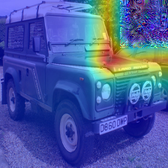}\\
Jeep & Doberman & Doberman & Doberman & Doberman \\
&&&&\\
\begin{sideways} \enspace \enspace \footnotesize Target: Centipede \end{sideways}
\includegraphics[width=.16\textwidth]{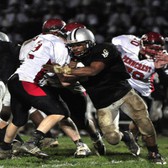}&
\includegraphics[width=.16\textwidth]{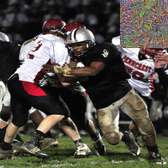}&
\includegraphics[width=.16\textwidth]{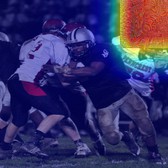}&
\includegraphics[width=.16\textwidth]{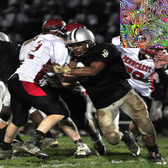}&
\includegraphics[width=.16\textwidth]{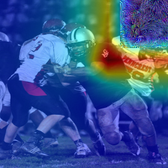}\\
Football Helmet & Centipede & Centipede & Centipede & Centipede \\
  \hline
\end{tabular}
  \caption{Similar to Figure \ref{fig_patch_nontarget1}, but for {\bf DenseNet-121} network.}
\label{fig_patch_nontarget2}
  \end{center}
\end{figure*}

\begin{figure*}[!ht]
  \begin{center}
  \begin{tabular}{| c c c c c c|}
\hline  \footnotesize Orig & \footnotesize Orig - GCAM &\footnotesize{PGD Adv } & \footnotesize{PGD Adv - GCAM} & \footnotesize{Our Adv} & \footnotesize{Our Adv - GCAM}\\
 \hline
&&&&&\\
\begin{sideways} \tiny Target: European gallinule \end{sideways}
\includegraphics[width=.14\textwidth]{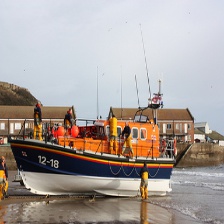}&
\includegraphics[width=.14\textwidth]{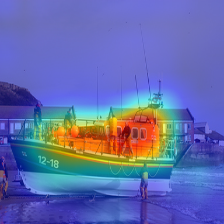}&
\includegraphics[width=.14\textwidth]{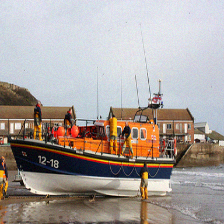}&
\includegraphics[width=.14\textwidth]{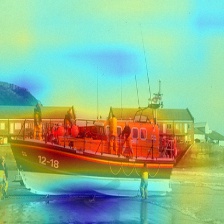}&
\includegraphics[width=.14\textwidth]{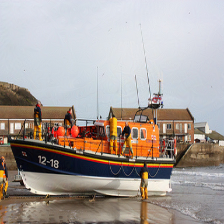}&
\includegraphics[width=.14\textwidth]{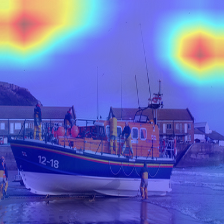}\\
\scriptsize Pred: Lifeboat & \scriptsize Grad-CAM ``Lifeboat'' & \scriptsize Pred: European & \scriptsize Grad-CAM ``Lifeboat'' & \scriptsize Pred: European & \scriptsize Grad-CAM ``Lifeboat'' \\
\vspace{-0.06in}
& & \scriptsize gallinule & & \scriptsize gallinule &\\
&&&&&\\

\begin{sideways} \enspace \scriptsize Target: Dowitcher \end{sideways}
\includegraphics[width=.14\textwidth]{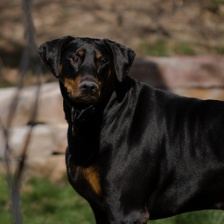}&
\includegraphics[width=.14\textwidth]{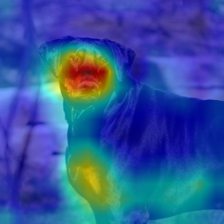}&
\includegraphics[width=.14\textwidth]{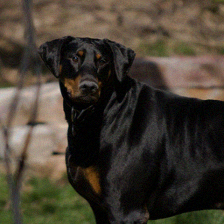}&
\includegraphics[width=.14\textwidth]{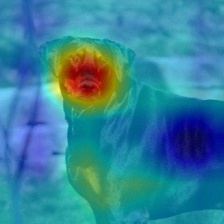}&
\includegraphics[width=.14\textwidth]{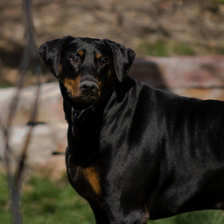}&
\includegraphics[width=.14\textwidth]{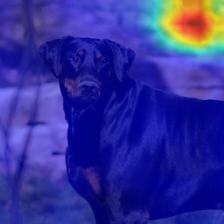}\\
\scriptsize Pred: Doberman & \scriptsize Grad-CAM ``Doberman'' & \scriptsize Pred: Dowitcher & \scriptsize Grad-CAM ``Doberman'' & \scriptsize Pred: Dowitcher & \scriptsize Grad-CAM ``Doberman'' \\
&&&&&\\

\begin{sideways} \enspace \enspace \scriptsize Target: Leonberg \end{sideways}
\includegraphics[width=.14\textwidth]{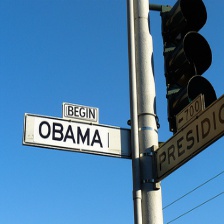}&
\includegraphics[width=.14\textwidth]{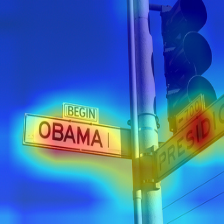}&
\includegraphics[width=.14\textwidth]{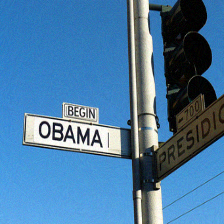}&
\includegraphics[width=.14\textwidth]{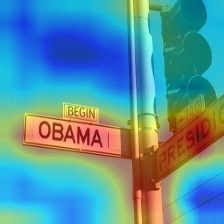}&
\includegraphics[width=.14\textwidth]{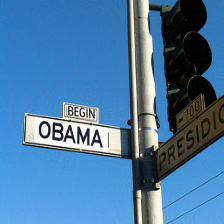}&
\includegraphics[width=.14\textwidth]{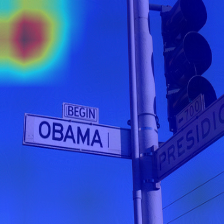}\\
\scriptsize Pred: Street Sign & \scriptsize Grad-CAM ``Street Sign'' & \scriptsize Pred: Leonberg & \scriptsize Grad-CAM ``Street Sign'' & \scriptsize Pred: Leonberg & \scriptsize Grad-CAM ``Street Sign'' \\
&&&&&\\

\begin{sideways} \enspace \quad \scriptsize Target: Potpie \end{sideways}
\includegraphics[width=.14\textwidth]{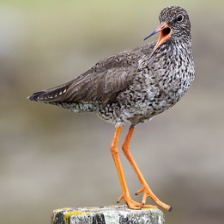}&
\includegraphics[width=.14\textwidth]{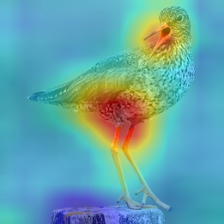}&
\includegraphics[width=.14\textwidth]{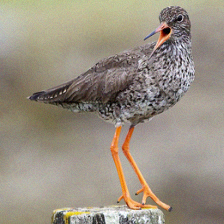}&
\includegraphics[width=.14\textwidth]{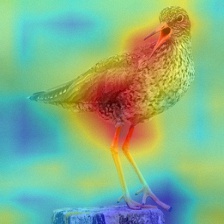}&
\includegraphics[width=.14\textwidth]{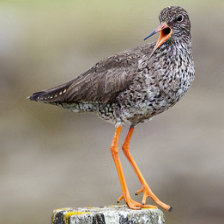}&
\includegraphics[width=.14\textwidth]{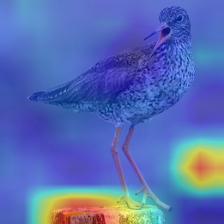}\\
\scriptsize Pred: Redshank & \scriptsize Grad-CAM ``Redshank'' & \scriptsize Pred: Potpie & \scriptsize Grad-CAM ``Redshank'' & \scriptsize Pred: Potpie & \scriptsize Grad-CAM ``Redshank'' \\

\hline
\end{tabular}
  \caption{\textbf{Targeted regular adversarial examples:} As described in Section \ref{exp_misleading_adv_examples}, we use an ImageNet pretrained VGG 19-BN network to perform a targeted attack using our method as well as using standard PGD method. Note that in this case, unlike other experiments, we compare Grad-CAM for the {\it{original}} category and not the target one. The predicted label is written under each image. The attack was successful for all images. Note that compared to the original image and the PGD adversarial image, the Grad-CAM for our adversarial image fires less on the object. This attack not only reduces the probability of the original category, but also changes its interpretation. Images are from ImageNet validation set. The quantitative results are in Table \ref{table:PGD}.}
\vspace{-.15in}
\label{fig_non_patch_target1}
  \end{center}
\end{figure*}

\begin{figure*}[!ht]
  \begin{center}
  \begin{tabular}{| c c c c c|}
    \hline
\footnotesize Original & \footnotesize Adv. Patch & \footnotesize Adv. Patch - GCAM & \footnotesize Our Patch & \footnotesize Our Patch - GCAM\\
  \hline
&&&&\\
\begin{sideways} \enspace \enspace \footnotesize Target: Aeroplane \end{sideways}
\includegraphics[width=.16\textwidth]{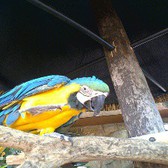}&
\includegraphics[width=.16\textwidth]{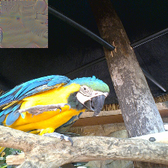}&
\includegraphics[width=.16\textwidth]{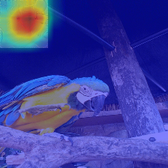}&
\includegraphics[width=.16\textwidth]{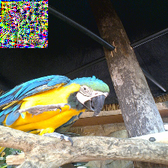}&
\includegraphics[width=.16\textwidth]{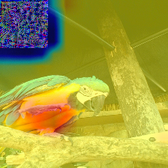}\\
Bird & Aeroplane & Aeroplane & Aeroplane & Aeroplane \\
&&&&\\
\begin{sideways} \qquad \footnotesize Target: Bike \end{sideways}
\includegraphics[width=.16\textwidth]{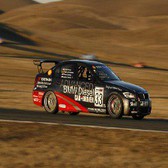}&
\includegraphics[width=.16\textwidth]{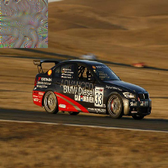}&
\includegraphics[width=.16\textwidth]{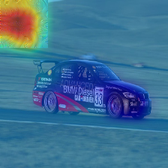}&
\includegraphics[width=.16\textwidth]{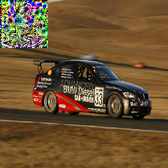}&
\includegraphics[width=.16\textwidth]{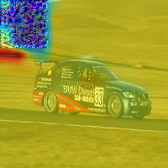}\\
Car & Bike & Bike & Bike & Bike \\
&&&&\\
\begin{sideways} \qquad \footnotesize Target: Bird \end{sideways}
\includegraphics[width=.16\textwidth]{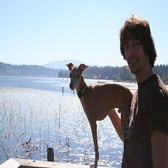}&
\includegraphics[width=.16\textwidth]{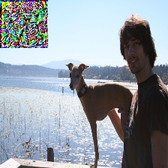}&
\includegraphics[width=.16\textwidth]{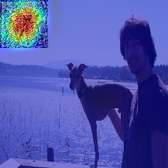}&
\includegraphics[width=.16\textwidth]{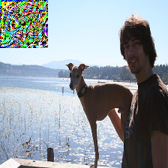}&
\includegraphics[width=.16\textwidth]{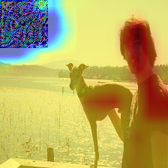}\\
Person & Bird & Bird & Bird & Bird \\
&&&&\\
\begin{sideways} \qquad \footnotesize Target: Boat \end{sideways}
\includegraphics[width=.16\textwidth]{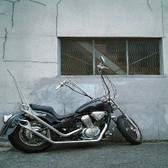}&
\includegraphics[width=.16\textwidth]{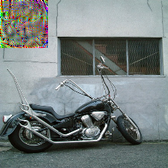}&
\includegraphics[width=.16\textwidth]{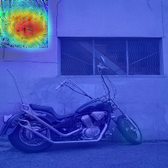}&
\includegraphics[width=.16\textwidth]{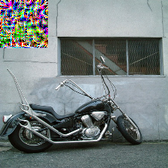}&
\includegraphics[width=.16\textwidth]{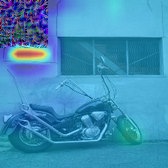}\\
Motorbike & Boat & Boat & Boat & Boat \\
  \hline
\end{tabular}
  \caption{{\bf Universal targeted patch attack:} As described in Section \ref{sec:universal}, we compare Grad-CAM of our universal attack on GAIN$_{ext}$ with the regular adversarial patch. The predicted label is written under each image, the targeted attack was successful for all images, and Grad-CAM is always computed for the target category. Note that each row shows the result for a different category chosen as the universal target. Images are from PASCAL VOC-2012 validation set. The quantitative results are in Table \ref{fig:comparison_patch_univ_heatmap_GAIN}.}
\label{fig_patch_univ1}
  \end{center}
\end{figure*}

\begin{figure*}[!ht]
  \begin{center}
  \begin{tabular}{| c c c c c|}
    \hline
\footnotesize Original & \footnotesize Adv. Patch & \footnotesize Adv. Patch - GCAM & \footnotesize Our Patch & \footnotesize Our Patch - GCAM\\
  \hline
&&&&\\
\begin{sideways} \qquad \footnotesize Target: Bottle \end{sideways}
\includegraphics[width=.16\textwidth]{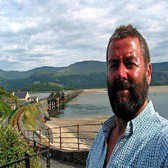}&
\includegraphics[width=.16\textwidth]{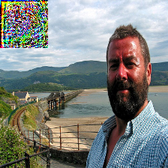}&
\includegraphics[width=.16\textwidth]{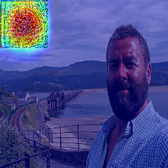}&
\includegraphics[width=.16\textwidth]{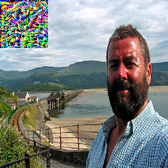}&
\includegraphics[width=.16\textwidth]{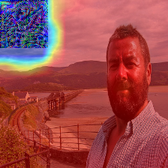}\\
Person & Bottle & Bottle & Bottle & Bottle \\
&&&&\\
\begin{sideways} \qquad \footnotesize Target: Bus \end{sideways}
\includegraphics[width=.16\textwidth]{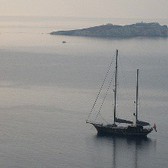}&
\includegraphics[width=.16\textwidth]{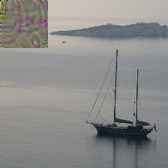}&
\includegraphics[width=.16\textwidth]{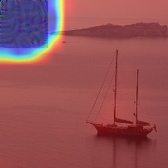}&
\includegraphics[width=.16\textwidth]{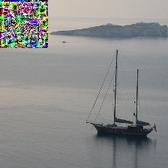}&
\includegraphics[width=.16\textwidth]{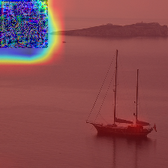}\\
Boat & Bus & Bus & Bus & Bus \\
&&&&\\
\begin{sideways} \qquad \footnotesize Target: Cat \end{sideways}
\includegraphics[width=.16\textwidth]{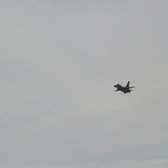}&
\includegraphics[width=.16\textwidth]{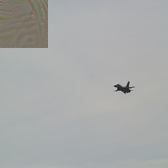}&
\includegraphics[width=.16\textwidth]{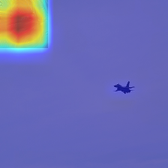}&
\includegraphics[width=.16\textwidth]{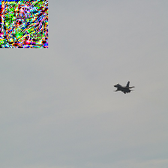}&
\includegraphics[width=.16\textwidth]{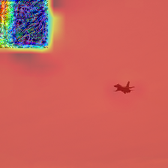}\\
Aeroplane & Cat & Cat & Cat & Cat \\
&&&&\\
\begin{sideways} \qquad \footnotesize Target: Car \end{sideways}
\includegraphics[width=.16\textwidth]{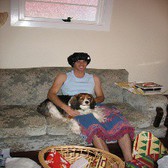}&
\includegraphics[width=.16\textwidth]{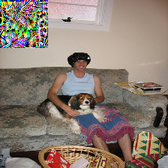}&
\includegraphics[width=.16\textwidth]{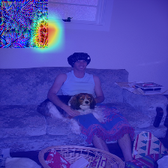}&
\includegraphics[width=.16\textwidth]{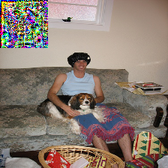}&
\includegraphics[width=.16\textwidth]{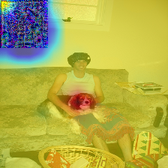}\\
Dog & Car & Car & Car & Car \\
  \hline
\end{tabular}
  \caption{Similar to Figure \ref{fig_patch_univ1}, but for different target categories. Interestingly, in the second row, regular adversarial patch is already hidden in Grad-CAM although it is not optimized for.}
\label{fig_patch_univ2}
  \end{center}
\end{figure*}

\begin{figure*}[!ht]
  \begin{center}
  \begin{tabular}{| c c c c c|}
    \hline
\footnotesize Original & \footnotesize Adv. Patch & \footnotesize Adv. Patch - GCAM & \footnotesize Our Patch & \footnotesize Our Patch - GCAM\\
  \hline
&&&&\\
\begin{sideways} \qquad \footnotesize Target: Chair \end{sideways}
\includegraphics[width=.16\textwidth]{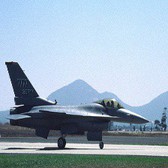}&
\includegraphics[width=.16\textwidth]{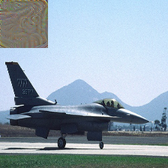}&
\includegraphics[width=.16\textwidth]{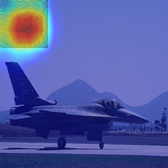}&
\includegraphics[width=.16\textwidth]{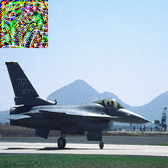}&
\includegraphics[width=.16\textwidth]{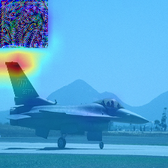}\\
Aeroplane & Chair & Chair & Chair & Chair \\
&&&&\\
\begin{sideways} \qquad \footnotesize Target: Cow \end{sideways}
\includegraphics[width=.16\textwidth]{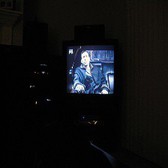}&
\includegraphics[width=.16\textwidth]{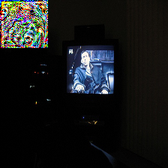}&
\includegraphics[width=.16\textwidth]{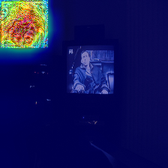}&
\includegraphics[width=.16\textwidth]{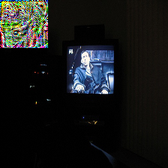}&
\includegraphics[width=.16\textwidth]{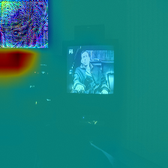}\\
TV/Monitor & Cow & Cow & Cow & Cow \\
&&&&\\
\begin{sideways} \enspace \footnotesize Target: Dining Table \end{sideways}
\includegraphics[width=.16\textwidth]{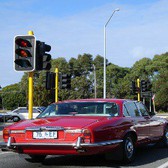}&
\includegraphics[width=.16\textwidth]{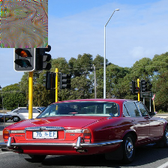}&
\includegraphics[width=.16\textwidth]{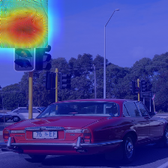}&
\includegraphics[width=.16\textwidth]{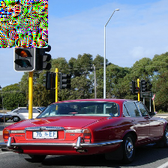}&
\includegraphics[width=.16\textwidth]{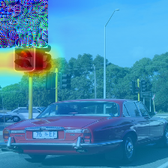}\\
Car & Dining Table & Dining Table & Dining Table & Dining Table \\
&&&&\\
\begin{sideways} \qquad \footnotesize Target: Dog \end{sideways}
\includegraphics[width=.16\textwidth]{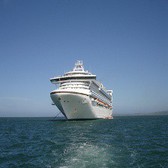}&
\includegraphics[width=.16\textwidth]{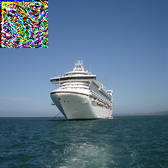}&
\includegraphics[width=.16\textwidth]{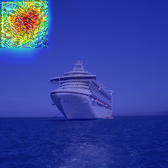}&
\includegraphics[width=.16\textwidth]{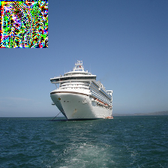}&
\includegraphics[width=.16\textwidth]{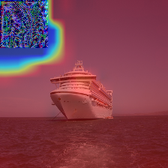}\\
Boat & Dog & Dog & Dog & Dog \\
  \hline
\end{tabular}
  \caption{Similar to Figure \ref{fig_patch_univ1} and \ref{fig_patch_univ2}, but for different target categories.}
\label{fig_patch_univ3}
  \end{center}
\end{figure*}

\end{document}